%% file: main.tex
\documentclass[runningheads]{llncs}
\usepackage{graphicx}

\usepackage{amsmath,amssymb} 
\usepackage{booktabs}
\usepackage{multirow}
\usepackage{array, makecell, rotating, boldline}
\usepackage{dblfloatfix}
\usepackage{setspace}
\usepackage{comment}

\usepackage[labelfont=bf,skip=2pt]{caption}
\usepackage{subcaption}
\usepackage{lipsum}
\DeclareCaptionFont{mytiny}{\fontsize{2pt}{3pt}\selectfont#1}

\usepackage{cite}

\usepackage{tikz}

\usepackage{xcolor}
\definecolor{darkgreen}{rgb}{0.0, 0.5, 0.0}

\newcommand{\splitcellscripttwo}[2]{\tiny\begin{tabular}{@{}c@{}}\scriptsize #1 \\ \scriptsize #2 \end{tabular}}

\usepackage{pifont}

\usepackage{epsfig}

\usepackage{makecell}
\newcommand{\hbline}{\Xhline{3.5\arrayrulewidth}}

\newcolumntype{?}{!{\vrule width 1.75pt}}

\newcommand{\dispskip}{1pt}
\usepackage[pagebackref=true,breaklinks=true,letterpaper=true,colorlinks,bookmarks=false]{hyperref}

\begin{document}

\title{Bidirectional Conditional\\Generative Adversarial Networks} 
\titlerunning{BiCoGAN} 


\author{Ayush Jaiswal \and
Wael AbdAlmageed \and
Yue Wu \and
Premkumar Natarajan}
%

\authorrunning{A.Jaiswal et al.} 


\institute{USC Information Sciences Institute, Marina del Rey, CA, USA\\
\email{\{ajaiswal, wamageed, yue\_wu, pnataraj\}@isi.edu}}

\maketitle

\begin{abstract}
Conditional Generative Adversarial Networks (cGANs) are generative models that can produce data samples ($x$) conditioned on both latent variables ($z$) and known auxiliary information ($c$). We propose the Bidirectional cGAN (BiCoGAN), which effectively disentangles $z$ and $c$ in the generation process and provides an encoder that learns inverse mappings from $x$ to both $z$ and $c$, trained jointly with the generator and the discriminator. We present crucial techniques for training BiCoGANs, which involve an extrinsic factor loss along with an associated dynamically-tuned importance weight. As compared to other encoder-based cGANs, BiCoGANs encode $c$ more accurately, and utilize $z$ and $c$ more effectively and in a more disentangled way to generate samples.
\end{abstract}


\input{introduction}


\input{related_work}


\input{preliminaries}


\input{bcgan}


\input{eval}


\section{Conclusions}
\label{sec:conclusion}

We presented the bidirectional conditional GAN framework that effectively generates data conditioned on intrinsic and extrinsic factors in a disentangled manner and provides a jointly trained encoder to encode data into \emph{both} intrinsic and extrinsic factors underlying the data distribution. We presented necessary techniques for training BiCoGANs that incorporate an extrinsic factor loss with an associated importance weight. We showed that BiCoGAN exhibits state-of-the-art performance at encoding extrinsic factors of data and at disentangling intrinsic and extrinsic factors during generation on MNIST and CelebA. We provided results on the Chairs dataset to show that it works well with continuous extrinsic factors also. Finally, we showed that state-of-the-art performance can be achieved at predicting previously unseen attributes using BiCoGAN embeddings, demonstrating that the encodings can be used for downstream tasks.
\newline\newline\noindent\textbf{Acknowledgements.}\quad This work is based on research sponsored by the Defense Advanced Research Projects Agency under agreement number FA8750-16-2-0204. The U.S. Government is authorized to reproduce and distribute reprints for governmental purposes  notwithstanding any copyright notation thereon. The views and conclusions contained herein are those of the authors and should not be interpreted as necessarily representing the official policies or endorsements, either expressed or implied, of the Defense Advanced Research Projects Agency or the U.S. Government.



\bibliographystyle{splncs04}
\bibliography{main}
\end{document}

%% file: introduction.tex
\section{Introduction}

Generative Adversarial Networks (GAN)~\cite{bib:gan} have recently gained immense popularity in generative modeling of data from complex distributions for a variety of applications such as image editing~\cite{bib:icgan}, image synthesis from text descriptions~\cite{bib:im_synth}, image super-resolution~\cite{bib:super_res}, video summarization~\cite{bib:video_summ}, and others~\cite{bib:gan_app_6,bib:gan_app_5,bib:gan_app_2,bib:gan_app_10,bib:gan_app_3,bib:unsupervised_aug,bib:gan_app_7,bib:gan_app_1,bib:gan_app_11}. 
GANs essentially learn a mapping from a latent distribution to a higher dimensional, more complex data distribution. Many variants of the GAN framework have been recently developed to augment GANs with more functionality and to improve their performance in both data modeling and target applications~\cite{bib:icgan,bib:bigan,bib:ali,bib:gan_variant_1,bib:gan_variant_2,bib:aae,bib:vae_gan,bib:avb,bib:cgan,bib:lr_gan}. Conditional GAN (cGAN)~\cite{bib:cgan} is a variant of standard GANs that was introduced to augment GANs with the capability of conditional generation of data samples based on both latent variables (or intrinsic factors) and known auxiliary information (or extrinsic factors) such as class information or associated data from other modalities. Desired properties of cGANs include the ability to disentangle the intrinsic and extrinsic factors, and also disentangle the components of extrinsic factors from each other, in the generation process, such that the incorporation of a factor minimally influences that of the others. Inversion of such a cGAN provides a disentangled information-rich representation of data, which can be used for downstream tasks (such as classification) instead of raw data. Therefore, an optimal framework would be one that ensures that the generation process uses factors in a disentangled manner \emph{and} provides an encoder to invert the generation process, giving us a disentangled encoding. The existing equivalent of such a framework is the Invertible cGAN (IcGAN)~\cite{bib:icgan}, which learns inverse mappings to intrinsic and extrinsic factors for \emph{pretrained} cGANs. The limitations of post-hoc training of encoders in IcGANs are that it prevents them from (1) influencing the disentanglement of factors during generation, and (2) learning the inverse mapping to intrinsic factors effectively, as noted for GANs in~\cite{bib:ali}. Other encoder-based cGAN models either do not encode extrinsic factors~\cite{bib:aae} or encode them in fixed-length continuous vectors that do not have an explicit form~\cite{bib:vae_gan}, which prevents the generation of data with arbitrary combinations of extrinsic attributes.

We propose the Bidirectional Conditional GAN (BiCoGAN), which overcomes the deficiencies of the aforementioned encoder-based cGANs. The encoder in the proposed BiCoGAN is trained \emph{simultaneously} with the generator and the discriminator, and learns inverse mappings of data samples to \emph{both intrinsic and extrinsic factors}. Hence, our model exhibits implicit regularization, mode coverage and robustness against mode collapse similar to Bidirectional GANs (BiGANs)~\cite{bib:bigan,bib:ali}. However, training BiCoGANs na\"ively does not produce good results in practice, because the encoder fails to model the inverse mapping to extrinsic attributes and the generator fails to incorporate the extrinsic factors while producing data samples. We present crucial techniques for training BiCoGANs, which address both of these problems. BiCoGANs outperform IcGANs on both encoding and generation tasks, and have the added advantages of end-to-end training, robustness to mode collapse and fewer model parameters. Additionally, the BiCoGAN-encoder outperforms IcGAN and the state-of-the-art methods on facial attribute prediction on cropped and aligned CelebA~\cite{bib:lnet_anet} images. Furthermore, state-of-the-art performance can be achieved at predicting previously unseen facial attributes using features learned by our model instead of images. The proposed model is significantly different from the conditional extension of the ALI model (cALIM)~\cite{bib:ali}. cALIM does not encode extrinsic attributes from data samples. It requires both data samples and extrinsic attributes as inputs to encode the intrinsic features. Thus, their extension is a conditional BiGAN, which is functionally different from the proposed bidirectional cGAN.

This paper has the following contributions. It (1) introduces the new BiCoGAN framework, (2) provides crucial techniques for training BiCoGANs, and (3) presents a thorough comparison of BiCoGANs with other encoder-based GANs, showing that our method achieves the state-of-the-art performance on several metrics. The rest of the paper is organized as follows. Section~\ref{sec:related_work} discusses related work. In Section~\ref{sec:preliminaries} we review the building blocks underlying the design of our model: GANs, cGANs and BiGANs. Section~\ref{sec:bcgan} describes our BiCoGAN framework and techniques for training BiCoGANs. Qualitative and quantitative analyses of our model are presented in Section~\ref{sec:analyses}. Section~\ref{sec:conclusion} concludes the paper and provides directions for future research.

%% file: related_work.tex
\section{Related Work}
\label{sec:related_work}

Perarnau et al.~\cite{bib:icgan} developed the IcGAN model to learn inverse mappings of a \emph{pretrained} cGAN from data samples to intrinsic and extrinsic attributes using two independent encoders trained \emph{post-hoc}, one for each task. In their experiments they showed that using a common encoder did not perform well. In contrast, the proposed BiCoGAN model incorporates a single encoder to embed both intrinsic and extrinsic factors, which is trained jointly with the generator and the discriminator from scratch.

BiGANs are related to autoencoders~\cite{bib:ae}, which also encode data samples and reconstruct data from compact embeddings. Donahue et al.~\cite{bib:bigan} show a detailed mathematical relationship between the two frameworks. Makhzani et al.~\cite{bib:aae} introduced an adversarial variant of autoencoders (AAE) that constrains the latent embedding to be close to a simple prior distribution (e.g., a multivariate Gaussian). Their model consists of an encoder $Enc$, a decoder $Dec$ and a discriminator. While the encoder and the decoder are trained with the reconstruction loss $\lVert x - Dec(Enc(x)) \rVert^{2}_{2}$ (where $x$ represents real data samples), the discriminator decides whether a latent vector comes from the prior distribution or from the encoder's output distribution. In their paper, they presented unsupervised, semi-supervised and supervised variants of AAEs. Supervised AAEs (SAAEs) have a similar setting as BiCoGANs. Both SAAE decoders and BiCoGAN generators transform intrinsic and extrinsic factors into data samples. However, SAAE encoders learn only intrinsic factors while encoders of the proposed BiCoGAN model learn both. While the structure of data samples is learned explicitly through the reconstruction loss in SAAE, it is learned implicitly in BiCoGANs.

Variational Autoencoders (VAE)~\cite{bib:vae} have also been trained adversarially in both unconditional and conditional settings~\cite{bib:vae_gan,bib:avb}. The conditional adversarial VAE of~\cite{bib:vae_gan} (cAVAE) encodes extrinsic factors of data into a fixed-length continuous vector $s$. This vector along with encoded latent attributes can be used to reconstruct images. However, $s$ is not interpretable and comes from encoding a real data sample. Hence, generating a new sample with certain desired extrinsic properties from a cAVAE requires first encoding a similar real data sample (with exactly those properties) to get its $s$. In comparison, such attributes can be explicitly provided to BiCoGANs for data generation.




%% file: preliminaries.tex
\section{Preliminaries}
\label{sec:preliminaries}

In this section, we introduce the mathematical notation and a brief description of the fundamental building blocks underlying the design of BiCoGANs including GANs, cGANs and BiGANs.

\subsection{Generative Adversarial Networks}

The working principle of the GAN framework is learning a mapping from a simple latent (or prior) distribution to the more complex data distribution. A GAN is composed of a generator and a discriminator. The goal of the generator is to produce samples that resemble real data samples, while the discriminator's objective is to differentiate between real samples and those generated by the generator. The data $x$ comes from the distribution $p_d$ and the latent vector $z$ is drawn from a prior distribution $p_z$. Therefore, the generator is a mapping $G(z; \theta_G)$ from $p_z$ to the generator's distribution $p_G$ with the goal of bringing $p_G$ as close as possible to $p_d$. On the other hand, the discriminator  $D(x; \theta_D)$ is simply a classifier that produces a scalar value $y \in [0, 1]$ indicating whether $x$ is from $p_G$ or from $p_d$. The generator and the discriminator play the minimax game (with the networks trained through backpropagation) as shown in Equation~\ref{eq:gan}.
\begingroup\setlength{\belowdisplayskip}{-15pt}
\begin{align}
\min_{G} \max_{D} V(D, G) = \mathbb{E}_{x \sim p_d(x)} \left [ \log D(x) \right ] + \mathbb{E}_{z \sim p_{z}(z)} \left [ \log (1 - D(G(z))) \right ]
\label{eq:gan}
\end{align}
\endgroup

\subsection{Conditional Generative Adversarial Networks}

Mirza et al.~\cite{bib:cgan} introduced conditional GAN (cGAN), which extends the GAN framework to the conditional setting where data can be generated conditioned on known auxiliary information such as class labels, object attributes, and associated data from different modalities. cGANs thus provide more control over the data generation process with an explicit way to communicate desired attributes of the data to be generated to the GAN. This can be thought of as using a new prior vector $\tilde{z}$ with two components $\tilde{z} = [z \ c]$, where $z$ represents latent \emph{intrinsic} factors and $c$ represents auxiliary \emph{extrinsic} factors. Hence, the generator is a mapping $G(\tilde{z}; \theta_G)$ from $p_{\tilde{z}}$ to $p_G$ and the discriminator models $D(x, c; \theta_D)$ that gives $y \in [0, 1]$. The cGAN discriminator also utilizes the knowledge of $c$ to determine if $x$ is real or fake. Thus, the generator must incorporate $c$ while producing $x$ in order to fool the discriminator. The model is trained with a similar minimax objective as the original GAN formulation, as shown in Equation~\ref{eq:cgan}.
\begingroup\setlength{\belowdisplayskip}{-15pt}
\begin{align}
\min_{G} \max_{D} V(D, G) = \mathbb{E}_{x \sim p_d(x)} \left [ \log D(x, c) \right ] + \mathbb{E}_{z \sim p_{\tilde{z}}(\tilde{z})} \left [ \log (1 - D(G(\tilde{z}), c)) \right ]
\label{eq:cgan}
\end{align}
\endgroup

\subsection{Bidirectional Generative Adversarial Networks}

The GAN framework provides a mapping from $z$ to $x$, but not another from $x$ to $z$. Such a mapping is highly useful as it provides an information-rich representation of $x$, which can be used as input for downstream tasks (such as classification) instead of the original data in simple yet effective ways~\cite{bib:bigan,bib:ali}. Donahue et al.~\cite{bib:bigan} and Dumoulin et al.~\cite{bib:ali} independently developed the BiGAN (or ALI) model that adds an encoder to the original generator-discriminator framework. The generator models the same mapping as the original GAN generator while the encoder is a  mapping $E(x; \theta_E)$ from $p_d$ to $p_E$ with the goal of bringing $p_E$ close to $p_z$. The discriminator is modified to incorporate both $z$ and $G(z)$ or both $x$ and $E(x)$ to make real/fake decisions as $D(z, G(z); \theta_D)$ or $D(E(x), x; \theta_D)$, respectively. Donahue et al.~\cite{bib:bigan} provide a detailed proof to show that under optimality, $G$ and $E$ must be inverses of each other to successfully fool the discriminator. The model is trained with the new minimax objective as shown in Equation~\ref{eq:bigan}.
\begin{align}
\min_{G, E} \max_{D} V(D, G, E) = \mathbb{E}_{x \sim p_d(x)}& \left [ \log D(E(x), x) \right ] \nonumber \\
&+ \mathbb{E}_{z \sim p_{z}(z)} \left [ \log (1 - D(z, G(z))) \right ]
\label{eq:bigan}
\end{align}

%% file: bcgan.tex
\section{Proposed Model --- Bidirectional Conditional GAN}
\label{sec:bcgan}

An optimal cGAN framework would be one in which (1) the extrinsic factors can be explicitly specified so as to enable data generation conditioned on arbitrary combinations of factors, (2) the generation process uses intrinsic and extrinsic factors in a disentangled manner, (3) the components of the extrinsic factors minimally affect each other while generating data, and (4) the generation process can be inverted, giving us a disentangled information-rich embedding of data. However, existing models fail to simultaneously fulfill all of these desired properties, as reflected in Table~\ref{tab:gan_properties}. Moreover, formulating and training such a cGAN model is difficult given the inherent complexity of training GANs and the added constraints required to achieve the said goals.

We design the proposed Bidirectional Conditional GAN (BiCoGAN) framework with the aforementioned properties as our foundational guidelines. While goal (1) is fulfilled by explicitly providing the extrinsic factors as inputs to the BiCoGAN generator, in order to accomplish goals (2) and (3), we design the BiCoGAN discriminator to check the consistency of the input data with the associated intrinsic and extrinsic factors. Thus, the BiCoGAN generator must effectively incorporate both the sets of factors into the generation process to successfully fool the discriminator. Finally, in order to achieve goal (4), we incorporate an encoder in the BiCoGAN framework that learns the inverse mapping of data samples to \emph{both} intrinsic and extrinsic factors. We train the encoder \emph{jointly} with the generator and discriminator to ascertain that it effectively learns the inverse mappings and improves the generation process through implicit regularization, better mode coverage and robustness against mode collapse (like BiGANs~\cite{bib:bigan,bib:ali}). Thus, BiCoGANs generate samples conditioned on desired extrinsic factors and \emph{effectively} encode real data samples into disentangled representations comprising \emph{both} intrinsic and extrinsic attributes. This provides an information-rich representation of data for auxiliary supervised semantic tasks~\cite{bib:bigan}, as well as a way for conditional data augmentation~\cite{bib:unsupervised_aug,bib:supervised_aug} to aid their learning. Figure~\ref{fig:bcgan} illustrates the proposed BiCoGAN framework.

\begingroup\setlength{\abovecaptionskip}{5pt}
\begin{figure}
\centering
\includegraphics[clip,width=0.825\linewidth]{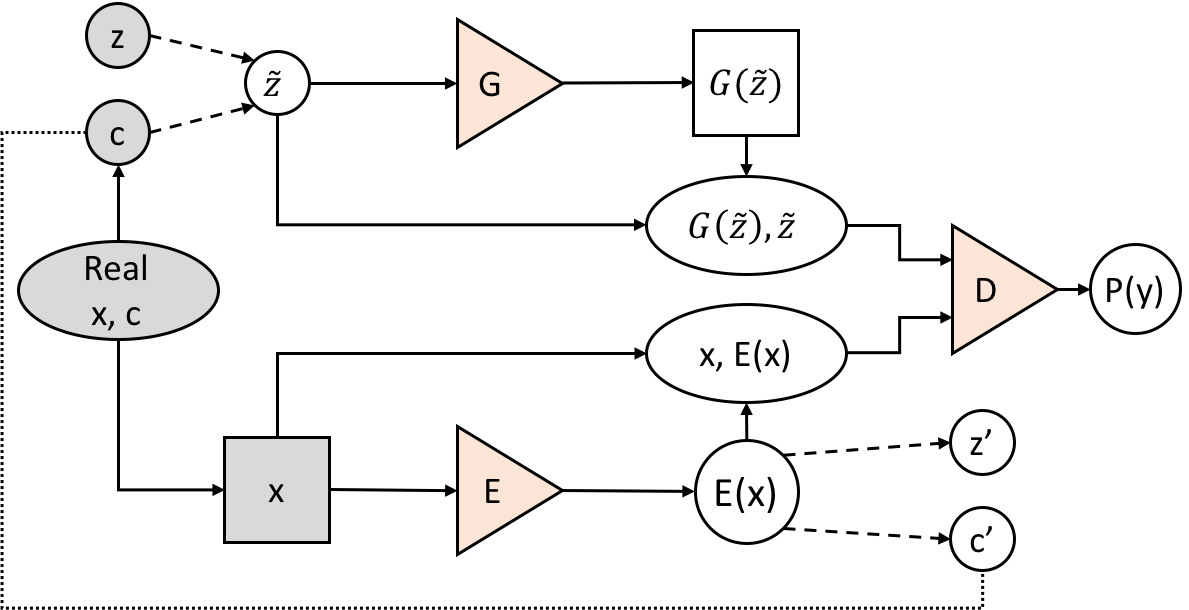}
\caption{Bidirectional Conditional Generative Adversarial Network. The dotted line indicates that $E$ is trained to predict the $c$ part of $E(x)$ with supervision.}
\label{fig:bcgan}
\end{figure}
\endgroup


The generator learns a mapping $G(\tilde{z}; \theta_G)$ from the distribution $p_{\tilde{z}}$ (where $\tilde{z} = [z \ c]$) to $p_G$ with the goal of bringing $p_G$ close to $p_{data}$ while the encoder models $E(x; \theta_E)$ from $p_{data}$ to $p_E$ with the goal of bringing $p_E$ close to $p_{\tilde{z}}$. The discriminator makes real/fake decisions as $D(\tilde{z}, G(\tilde{z}); \theta_D)$ or $D(E(x), x; \theta_D)$. It is important to note that the proposed BiCoGAN encoder must learn the inverse mapping of $x$ to $z$ \emph{and} $c$ just like the generator must learn to incorporate \emph{both} into the generation of data samples in order to fool the discriminator, following from the invertibility under optimality theorem of BiGANs~\cite{bib:bigan,bib:ali}. However, in practice, such optimality is difficult to achieve, especially when the prior vector contains structured information or has a complex distribution. While the intrinsic factors are sampled randomly from a simple latent distribution, the extrinsic factors are much more specialized and model specific forms of high-level information, such as class labels or object attributes, making their underlying distribution significantly more difficult to model. To address this challenge, we introduce the \textbf{\textit{extrinsic factor loss}} (EFL) as an explicit mechanism that helps guide BiCoGANs to better encode extrinsic factors. This is built on the fact that the $c$ associated with each real data sample is known during training, and can, thus, be used to improve the learning of inverse mappings from $x$ to $c$. \emph{We do not give an explicit form to EFL in the BiCoGAN objective because the choice of the loss function depends on the nature of $c$, and hence, on the dataset/domain.}

Adding EFL to the BiCoGAN objective is not sufficient to achieve the best results for both encoding $c$ and generating $x$ that incorporates the knowledge of $c$. This is justified by the fact that the training process has no information about the inherent difficulty of encoding $c$ (specific to the domain). Thus, it is possible that the backpropagated gradients of the EFL (to the encoder) are distorted by those from the discriminator in the BiCoGAN framework. Therefore, we multiply EFL with an importance weight, which we denote by $\gamma$ and refer to as the EFL weight (EFLW), in the BiCoGAN objective as shown in Equation~\ref{eq:bcgan}.
\begin{align}
\min_{G, E}& \max_{D} V(D, G, E) = \mathbb{E}_{x \sim p_{data}(x)} \left [ \log D(E(x), x) \right ] \nonumber \\
&+ \gamma \ \mathbb{E}_{(x, c) \sim p_{data}(x, c)} \left [ \text{EFL}(c, E_{c}(x)) \right ] + \ \ \mathbb{E}_{z \sim p_{\tilde{z}}(\tilde{z})} \left [ \log (1 - D(\tilde{z}, G(\tilde{z}))) \right ]
\label{eq:bcgan}
\end{align}


The importance weight $\gamma$ can be chosen as a constant value or a dynamic parameter that keeps changing during training to control the focus of the training between the na\"ive adversarial objective and the EFL. While the former option is straightforward, the latter requires some understanding of the dynamics between the original generator-discriminator setup of cGANs and the additional encoder as introduced in the proposed BiCoGAN model. It can be seen that the objective of the generator is significantly more difficult than that of the encoder, making the former more vulnerable to instability during training. Thus, in the dynamic setting, we design $\gamma$ as a \emph{clipped} exponentially increasing variable that starts with a small initial value, i.e., $\gamma = \min (\alpha e^{\rho t}, \phi)$, where $\alpha$ is the initial value for $\gamma$, $\phi$ is its maximum value, $\rho$ controls the rate of exponential increase and $t$ indicates the number of epochs the model has already been trained. This is motivated by a similar approach introduced in~\cite{bib:mtl_evolve} for deep multi-task learning.

%% file: eval.tex
\section{Experimental Evaluation}
\label{sec:analyses}
We evaluate the performance of the encoder and the generator of the proposed BiCoGAN model jointly and independently, and compare it with other encoder-based GANs, specifically, IcGAN, cALIM and cAVAE on various tasks. We also evaluate the effect of EFL and EFLW on BiCoGAN training.

\subsection{Datasets}

All  models are evaluated on the MNIST~\cite{bib:mnist} handwritten digits dataset and the CelebA~\cite{bib:lnet_anet} dataset of celebrity faces with annotated facial attributes. We consider the class labels in the MNIST dataset as extrinsic factors and components of writing styles as intrinsic factors. We select the same $18$ visually impactful facial attributes of the CelebA dataset as~\cite{bib:icgan} as extrinsic factors and all other factors of variation as intrinsic features. We did not evaluate the other GAN models on datasets for which their official implementations were not available. Therefore, we compare BiCoGAN with IcGAN and cAVAE on MNIST, and with IcGAN and cALIM on CelebA. We also present qualitative results of the proposed BiCoGAN model on the Chairs dataset~\cite{bib:chairs}. Each chair is rendered at $31$ different yaw angles, and cropped and downsampled to $32 \times 32$ dimensions. We use the yaw angle, a continuous value, as the extrinsic attribute for this dataset and all other factors of variation as intrinsic variables.



\subsection{Metrics}\label{sec:metrics}

We quantify the performance of encoding the extrinsic factors, $c$, using both mean accuracy ($A_c$) and mean $F_1$-score ($F_c$). We follow the approach in~\cite{bib:improved_gan} of using an external discriminative model to assess the quality of generated images. The core idea behind this approach is that the performance of an external model trained on real data samples should be similar when evaluated on both real and GAN-generated test samples. We trained a digit classifier using a simple convolutional neural network for MNIST~\footnote{https://github.com/fchollet/keras/blob/master/examples/mnist\_cnn.py} and the attribute predictor Anet~\cite{bib:lnet_anet} model for CelebA. Thus, in our experimental settings, this metric also measures the ability of the generator in incorporating $c$ in the generation of $x$. We use both accuracy ($A^{Ext}_{gen}$) and $F_1$-score ($F^{Ext}_{gen}$) to quantify the performance of the external model. We show the accuracy and the $F_1$-score of these external models on real test datasets for reference as $A^{Ext}_{real}$ and $F^{Ext}_{real}$.  We also calculate the adversarial accuracy (AA) as proposed in~\cite{bib:lr_gan}. AA is calculated by training the external classifier on samples generated by a GAN and testing on real data. If the generator generalizes well and produces good quality images, the AA score should be similar to the $A^{Ext}_{gen}$ score. In order to calculate $A^{Ext}_{gen}$, $F^{Ext}_{gen}$ and AA, we use each GAN to generate a set of images $X_{gen}$. Denoting the real training dataset as $\langle X_{train}, C_{train} \rangle$, each image in $X_{gen}$ is created using a $c \in C_{train}$ combined with a randomly sampled $z$. $X_{gen}$ is then used as the testing set for calculating $A^{Ext}_{gen}$ and $F^{Ext}_{gen}$, and as the training set for calculating AA. Furthermore, we evaluate the ability of the GAN models to disentangle intrinsic factors from extrinsic attributes in the data generation process on the CelebA dataset using an identity-matching score (IMS). The motivation behind this metric is that the identity of generated faces should not change when identity-independent attributes (like hair color or the presence of eyeglasses) change. We first randomly generate $1000$ faces with ``male'' and ``black hair'' attributes and another $1000$ with ``female'' and ``black hair'' attributes. We then generate eight variations of these base images with the attributes: ``bangs'', ``receding hairline'', ``blond hair'', ``brown hair'', ``gray hair'', ``heavy makeup'', ``eyeglasses'' and ``smiling'' respectively. We encode all the generated images using a pretrained VGG-Face~\cite{bib:vgg_face} model. IMS is then calculated as the mean cosine similarity of the base images with their variations. We provide results on MNIST and CelebA for two settings of BiCoGANs; one where we prioritize the performance of the generator (BiCoGAN-gen) and another for that of the encoder (BiCoGAN-enc), which gives us an empirical upper bound on the performance of BiCoGAN encoders.

\subsection{Importance of Extrinsic Factor Loss}

\begin{figure}
\captionsetup[subfigure]{font=scriptsize,labelfont=scriptsize,aboveskip=2pt,belowskip=1pt}
\centering
\begin{subfigure}{0.243\textwidth}
\centering
\includegraphics[width=0.95\textwidth]{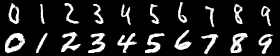}
\caption{}
\label{fig:bcgan_efl_mnist_qual_rand}
\end{subfigure}
\begin{subfigure}{0.243\textwidth}
\centering
\includegraphics[width=0.95\textwidth]{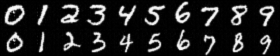}
\caption{}
\label{fig:icgan_mnist_qual_rand}
\end{subfigure}
\begin{subfigure}{0.243\textwidth}
\centering
\includegraphics[width=0.95\textwidth]{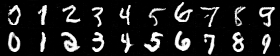}
\caption{}
\label{fig:cavae_mnist_qual_rand}
\end{subfigure}
\begin{subfigure}{0.243\textwidth}
\centering
\includegraphics[width=0.95\textwidth]{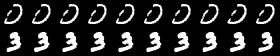}
\caption{}
\label{fig:bcgan_noefl_mnist_qual_rand}
\end{subfigure}
\caption{Randomly generated (MNIST) digits using (a) BiCoGAN with EFL, (b) IcGAN, (c) cAVAE and (d) BiCoGAN without EFL.}
\label{fig:bcgan_mnist_qual_rand}
\end{figure}

\begin{figure}
\captionsetup[subfigure]{font=scriptsize,labelfont=scriptsize,aboveskip=3pt,belowskip=1pt}
\centering
\begin{subfigure}{.45\textwidth}
\centering
\includegraphics[trim={0 0 0 0.39cm}, clip, width=0.825\textwidth]{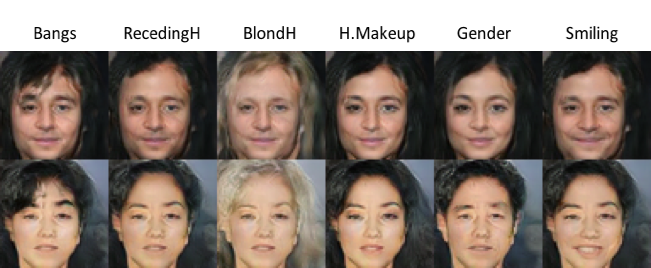}
\caption{BiCoGAN with EFL}
\label{fig:bcgan_efl_celeba_qual_rand}
\end{subfigure}
\begin{subfigure}{.45\textwidth}
\centering
\includegraphics[trim={0 0 0 0.39cm}, clip, width=0.825\textwidth]{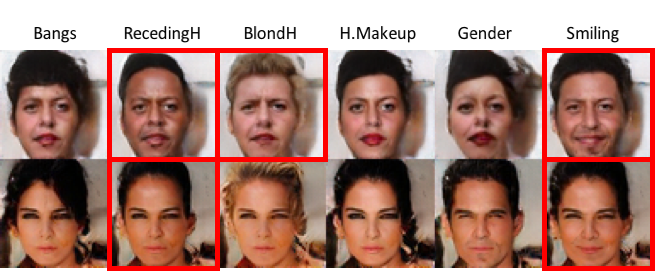}
\caption{cALIM}
\label{fig:calim_celeba_qual_rand}
\end{subfigure}
\begin{subfigure}{.45\textwidth}
\centering
\includegraphics[trim={0 0 0 0.39cm}, clip, width=0.825\textwidth]{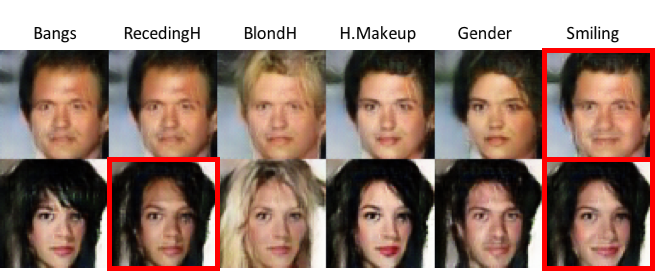}
\caption{IcGAN}
\label{fig:icgan_celeba_qual_rand}
\end{subfigure}
\begin{subfigure}{.45\textwidth}
\centering
\includegraphics[trim={0 0 0 0.39cm}, clip, width=0.825\textwidth]{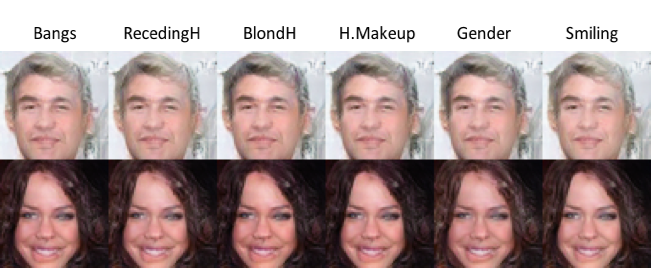}
\caption{BiCoGAN without EFL}
\label{fig:bcgan_noefl_celeba_qual_rand}
\end{subfigure}
\caption{Randomly generated (CelebA) faces. Base images are generated with black hair and gender as male (first row) and female (second row). ``Gender'' column indicates gender change. Red boxes show cases where unspecified attributes or latent factors are \emph{mistakenly} changed during generation.}
\label{fig:bcgan_celeba_qual_rand}
\end{figure}

\begin{figure}
\captionsetup[subfigure]{font=scriptsize,labelfont=scriptsize,aboveskip=3pt,belowskip=0.5pt}
\centering
\begin{subfigure}{.24\textwidth}
\centering
\includegraphics[width=\textwidth]{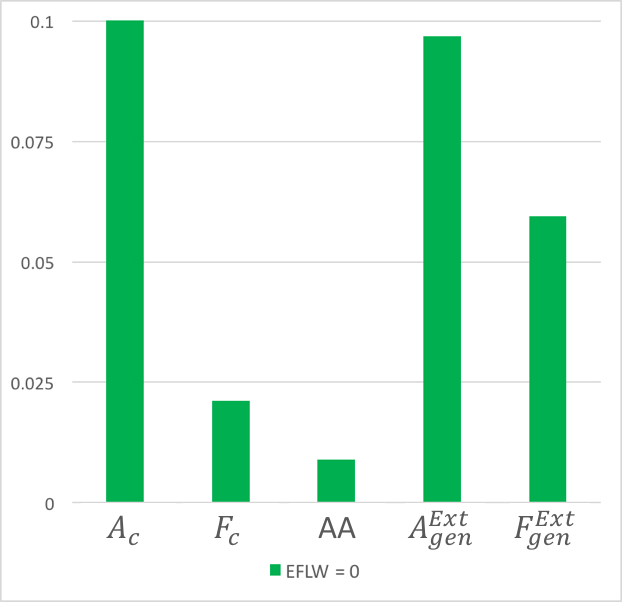}
\caption{}
\label{fig:bcgan_noefl_mnist_plot}
\end{subfigure}
\begin{subfigure}{.24\textwidth}
\centering
\includegraphics[width=\textwidth]{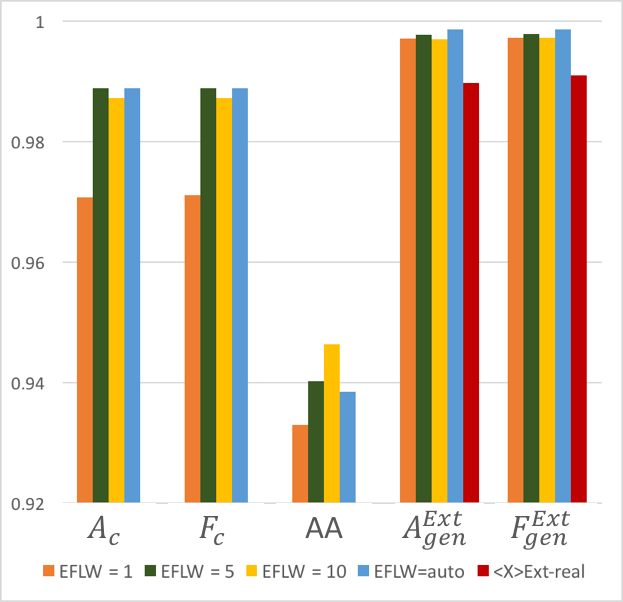}
\caption{}
\label{fig:bcgan_efl_mnist_plot}
\end{subfigure}
\begin{subfigure}{.24\textwidth}
\centering
\includegraphics[width=\textwidth]{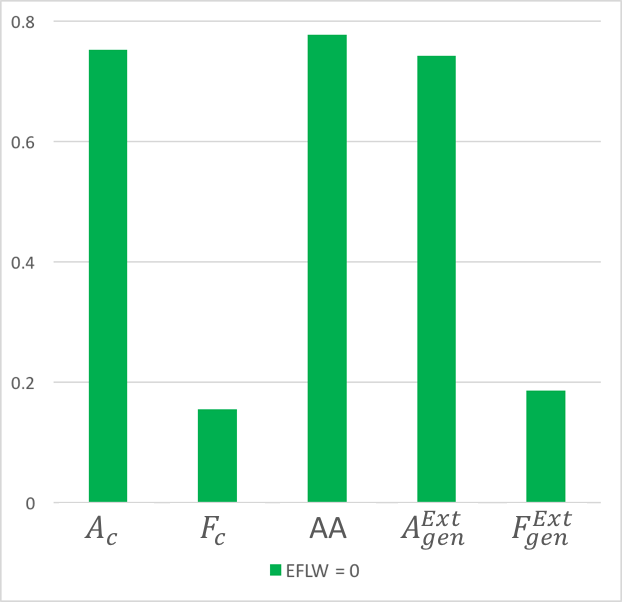}
\caption{}
\label{fig:bcgan_noefl_celeba_plot}
\end{subfigure}
\begin{subfigure}{.24\textwidth}
\centering
\includegraphics[width=\textwidth]{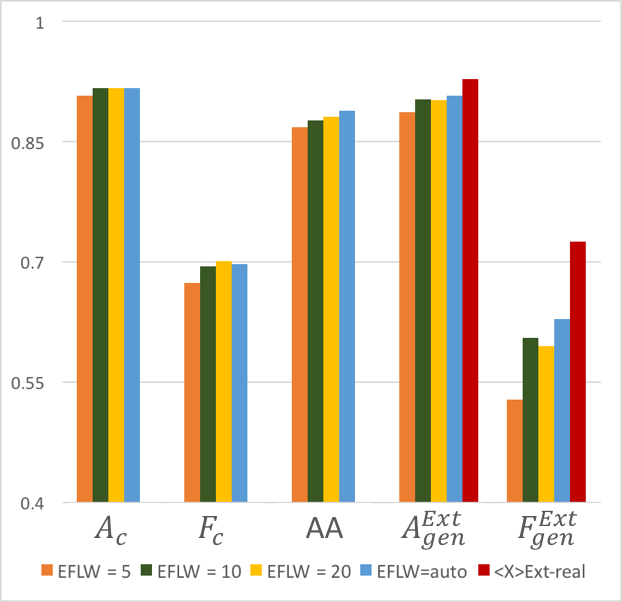}
\caption{}
\label{fig:bcgan_efl_celeba_plot}
\end{subfigure}
\caption{Influence of $\gamma$ for BiCoGANs trained on (a \& b) MNIST and on (c \& d) CelebA. $A_c$ \& $F_c$ show the performance of encoding $c$ while $A^{Ext}_{gen}$ \& $F^{Ext}_{gen}$ show that of data generation. ``EFLW=auto'' denotes the dynamic-$\gamma$ setting. The $A^{Ext}_{real}$ and $F^{Ext}_{real}$ values are shown as ``$\langle$X$\rangle$Ext-real'' values. The $Y$-axes of the plots have been scaled to easily observe differences.}
\label{fig:quant_gamma}
\end{figure}

We analyze the importance of incorporating EFL for training BiCoGAN and the influence of EFLW on its performance. Figures~\ref{fig:bcgan_noefl_mnist_qual_rand} and~\ref{fig:bcgan_noefl_celeba_qual_rand} show some examples of images randomly generated using a BiCoGAN trained without EFL on both MNIST and CelebA, respectively. We see that BiCoGANs are not able to incorporate $c$ into the data generation process when trained without EFL. The metrics discussed in Section \ref{sec:metrics} are calculated for BiCoGANs trained with $\gamma \in \{0, 1, 5, 10\}$ on MNIST, with $\gamma \in \{0, 5, 10, 20\}$ on CelebA, and with the dynamic setting of $\gamma = \min (\alpha e^{\rho t}, \phi)$, for $\alpha = 5$, $\rho = 0.25$ and $\phi = 10$, on both. Figure~\ref{fig:quant_gamma} summarizes our results. As before, we see that BiCoGANs are unable to learn the inverse mapping of $x$ to $c$ with $\gamma = 0$. The results show that increasing $\gamma$ up until a tipping point helps train BiCoGANs better. However, beyond that point, the EFL term starts dominating the overall objective, leading to degrading performance in the quality of generated images (as reflected by $A^{Ext}_{gen}$ and $F^{Ext}_{gen}$ scores). Meanwhile, the dynamic setting of $\gamma$ achieves the best results on both the datasets on almost all metrics, establishing its effectiveness at training BiCoGANs. It is also important to note that a dynamic $\gamma$ saves significant time and effort involved in selecting a constant $\gamma$ through manual optimization, which also depends on the complexity of the dataset. Therefore, we use BiCoGANs trained with dynamic $\gamma$ for the comparative results in the following sections.

\subsection{Conditional Generation}
\label{sec:cond_gen}

In this section, we evaluate the ability of the BiCoGAN generator to (1) generalize over the prior distribution of intrinsic factors, i.e., be able to generate images with random intrinsic factors, (2) incorporate extrinsic factors while producing images, and (3) disentangle intrinsic and extrinsic factors during generation.

Figures~\ref{fig:bcgan_efl_mnist_qual_rand},~\ref{fig:icgan_mnist_qual_rand}~and~\ref{fig:cavae_mnist_qual_rand} show some generated MNIST images with BiCoGAN, IcGAN and cAVAE, respectively. For each of these, we sampled $z$ vectors from the latent distribution (fixed along rows) and combined them with the digit class $c$ (fixed along columns). In order to vary $c$ for cAVAE, we picked a random image from each class and passed it through the cAVAE $s$-encoder to get its $s$-representation. This is required because $s$ in cAVAE does not have an explicit form and is instead a fixed-length continuous vector. The visual quality of the generated digits is similar for all the models with cAVAE producing slightly unrealistic images. Figures~\ref{fig:bcgan_efl_celeba_qual_rand},~\ref{fig:calim_celeba_qual_rand},~and~\ref{fig:icgan_celeba_qual_rand} show some generated CelebA images with BiCoGAN, cALIM and IcGAN respectively. For each row, we sampled $z$ from the latent distribution. We set $c$ to \texttt{male} and \texttt{black-hair} for the first row and \texttt{female} and \texttt{black-hair} for the second row. We then generate each image in the grids based on the combination of these with the new feature specified as the column header. The figures show that BiCoGANs perform the best at preserving intrinsic (like subject identity and lighting) and extrinsic factors (besides the specified new attribute). Hence, BiCoGAN outperforms the other models in disentangling the influence of $z$ and the components of $c$ in data generation.

\begin{table}
\centering
\setlength\tabcolsep{6pt}
\caption{Encoding and Generation Performance - MNIST}
\label{tab:quant_mnist}
\begin{tabular}{ c ? c | c ? c | c | c }
  \hbline
  \textbf{Model} & \textbf{$\boldsymbol{A_c}$} & \textbf{$\boldsymbol{F_c}$} & \textbf{AA} & \splitcellscripttwo{\textbf{$\boldsymbol{A^{Ext}_{gen}}$}}{(0.9897)} & \splitcellscripttwo{\textbf{$\boldsymbol{F^{Ext}_{gen}}$}}{(0.9910)} \\ \hbline
  cAVAE & -- & -- & \textbf{0.9614} & 0.8880 & 0.9910 \\
  IcGAN & 0.9871 & 0.9853 & 0.9360 & 0.9976  & \textbf{0.9986} \\ \hbline
  BiCoGAN-gen & 0.9888 & 0.9888 & 0.9384 & \textbf{0.9986} & \textbf{0.9986} \\ 
  BiCoGAN-enc & \textbf{0.9902} & \textbf{0.9906} & 0.9351 & 0.9933 & 0.9937 \\ \hbline
\end{tabular}
\end{table}

\begin{table}
\centering
\setlength\tabcolsep{4pt}
\caption{Encoding and Generation Performance - CelebA}
\label{tab:quant_celeba}
\begin{tabular}{ c ? c | c ? c | c | c ? c }
  \hbline
  \textbf{Model} & \textbf{$\boldsymbol{A_c}$} & \textbf{$\boldsymbol{F_c}$} & \textbf{AA} & \splitcellscripttwo{\textbf{$\boldsymbol{A^{Ext}_{gen}}$}}{(0.9279)} & \splitcellscripttwo{\textbf{$\boldsymbol{F^{Ext}_{gen}}$}}{(0.7253)} & \textbf{IMS}\\ \hbline
  cALIM & -- & -- & 0.9138 & \textbf{0.9139} & \textbf{0.6423} & 0.9085 \\
  IcGAN & 0.9127 & 0.5593 & 0.8760 & 0.9030 & 0.5969 & 0.8522 \\ \hbline 
  BiCoGAN-gen & 0.9166 & 0.6978 & \textbf{0.9174} & 0.9072 & 0.6289 & \textbf{0.9336} \\  
  BiCoGAN-enc & \textbf{0.9274} & \textbf{0.7338} & 0.8747 & 0.8849 & 0.5443 & 0.9286 \\ \hbline
\end{tabular}
\end{table}

We quantify the generation performance using $A^{Ext}_{gen}$, $F^{Ext}_{gen}$, AA and IMS. Table~\ref{tab:quant_mnist} shows results on MNIST for BiCoGAN, IcGAN and cAVAE. We show $A^{Ext}_{real}$ and $F^{Ext}_{real}$ for reference within parentheses in the $A^{Ext}_{gen}$ and $F^{Ext}_{gen}$ column headings, respectively. While BiCoGAN performs the best on $A^{Ext}_{gen}$ and $F^{Ext}_{gen}$ scores, cAVAE performs better on AA. This indicates that cAVAE is more prone to producing digits of wrong but easily confusable classes. Table~\ref{tab:quant_celeba} shows results on CelebA for BiCoGAN, IcGAN and cALIM. BiCoGAN outperforms IcGAN on almost all metrics. However, cALIM performs the best on $A^{Ext}_{gen}$ and $F^{Ext}_{gen}$. While this indicates that cALIM is better able to incorporate extrinsic factors for generating images, IMS indicates that cALIM does this at the cost of intrinsic factors. cALIM fails to effectively use the identity information contained in the intrinsic factors and disentangling it from the extrinsic attributes while generating images. BiCoGAN performs the best on IMS. BiCoGAN also performs the best on AA, indicating that it successfully generates diverse but realistic images.



\subsection{Encoding Extrinsic Factors}

We assess the performance of the models at encoding the extrinsic factors from data samples using the $A_c$ and $F_c$ metrics. We calculate these scores directly on the testing split of each dataset. Tables ~\ref{tab:quant_mnist} and ~\ref{tab:quant_celeba} show the performance of IcGAN and BiCoGAN in encoding $c$ on MNIST and CelebA, respectively. We note here that we cannot calculate $A_c$ and $F_c$ scores for cALIM because it does not encode $c$ from $x$ and for cAVAE because the $s$ it encodes does not have an explicit form. BiCoGAN consistently outperforms IcGAN at encoding extrinsic factors from data. Furthermore, we provide an attribute-level breakdown of accuracies for the CelebA dataset in Table~\ref{tab:attr_pred} and compare it with two state-of-the-art methods for cropped and aligned CelebA facial attribute prediction as reported in~\cite{bib:walk_learn}, namely, LNet+Anet~\cite{bib:lnet_anet} and WalkLearn~\cite{bib:walk_learn}. BiCoGAN outperforms the state-of-the-art methods even though the EFL directly responsible for it is only one part of the entire adversarial objective. This indicates that supervised tasks (like attribute prediction) can benefit from training the predictor with a generator and a discriminator in an adversarial framework like ours.

\begin{table}
\centering
\setlength\tabcolsep{5pt}
\caption{Attribute-level Breakdown of Encoder Accuracy - CelebA}
\label{tab:attr_pred}
\begin{tabular}{ c ? c | c ? c | c }
  \hbline
  \textbf{Attribute} & \textbf{LNet+ANet} & \textbf{WalkLearn} & \textbf{IcGAN} & \textbf{Ours} \\ \hbline
  Bald & 0.98 & 0.92 & \textbf{0.98} & \textbf{0.98} \\ 
  Bangs & 0.95 & \textbf{0.96} & 0.92 & 0.95 \\ 
  Black\_Hair & \textbf{0.88} & 0.84 & 0.83 & \textbf{0.88} \\ 
  Blond\_Hair & \textbf{0.95} & 0.92 & 0.93 & \textbf{0.95} \\ 
  Brown\_Hair & 0.8 & 0.81 & \textbf{0.87} & \textbf{0.87} \\ 
  Bushy\_Eyebrows & 0.9 & \textbf{0.93} & 0.91 & 0.92 \\ 
  Eyeglasses & \textbf{0.99} & 0.97 & 0.98 & \textbf{0.99} \\ 
  Gray\_Hair & 0.97 & 0.95 & \textbf{0.98} & \textbf{0.98} \\ 
  Heavy\_Makeup & 0.9 & \textbf{0.96} & 0.88 & 0.90 \\ 
  Male & \textbf{0.98} & 0.96 & 0.96 & 0.97 \\ 
  Mouth\_Slightly\_Open & 0.93 & \textbf{0.97} & 0.90 & 0.93 \\ 
  Mustache & 0.95 & 0.90 & \textbf{0.96} & \textbf{0.96} \\ 
  Pale\_Skin & 0.91 & 0.85 & 0.96 & \textbf{0.97} \\ 
  Receding\_Hairline & 0.89 & 0.84 & 0.92 & \textbf{0.93} \\ 
  Smiling & 0.92 & \textbf{0.98} & 0.90 & 0.92 \\ 
  Straight\_Hair & 0.73 & 0.75 & \textbf{0.80} & \textbf{0.80} \\ 
  Wavy\_Hair & 0.8 & \textbf{0.85} & 0.76 & 0.79 \\ 
  Wearing\_Hat & \textbf{0.99} & 0.96 & 0.98 & 0.98 \\ \hbline 
  MEAN & 0.91 & 0.91 & 0.91 & \textbf{0.93} \\ \hbline
\end{tabular}
\end{table}

\subsection{Image Reconstruction with Variations}

\begin{figure}
\captionsetup[subfigure]{font=scriptsize,labelfont=scriptsize,aboveskip=3pt, belowskip=-0.5pt}
\centering
\begin{subfigure}{.32\textwidth}
\centering
\includegraphics[trim={0 0 0 0.39cm}, clip, width=0.9\textwidth]{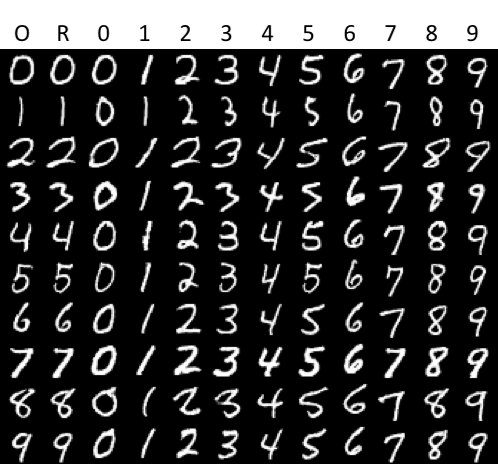}
\caption{BiCoGAN}
\label{fig:bcgan_efl_mnist_qual_autovar}
\end{subfigure}
\begin{subfigure}{.32\textwidth}
\centering
\includegraphics[trim={0 0 0 0.39cm}, clip, width=0.9\textwidth]{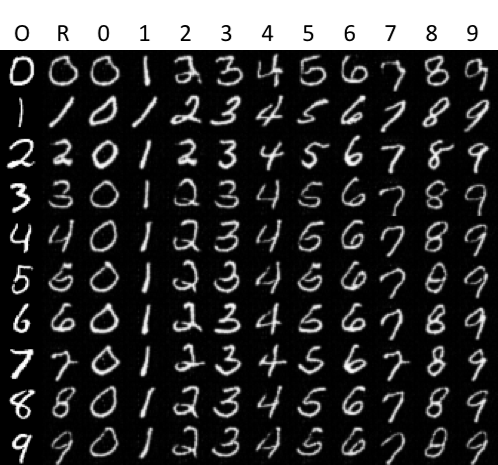}
\caption{IcGAN}
\label{fig:icgan_mnist_qual_autovar}
\end{subfigure}
\begin{subfigure}{.32\textwidth}
\centering
\includegraphics[trim={0 0 0 0.39cm}, clip, width=0.9\textwidth]{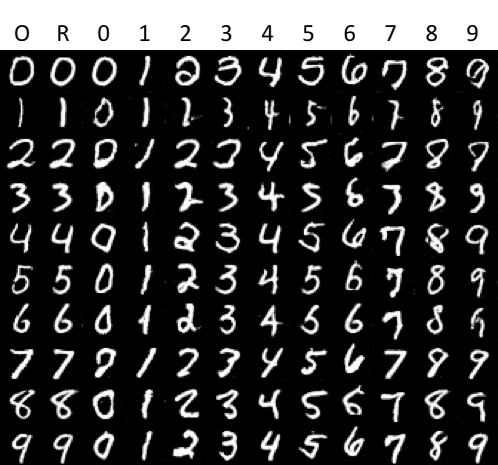}
\caption{cAVAE}
\label{fig:cavae_mnist_qual_autovar}
\end{subfigure}
\caption{MNIST images reconstructed with varied class information. Column ``O'' shows the real image; ``R'' shows the reconstruction. The following columns show images with  same $z$ but varied $c$.}
\label{fig:bcgan_mnist_qual_autovar}
\end{figure}

\begin{figure}
\captionsetup[subfigure]{font=scriptsize,labelfont=scriptsize,aboveskip=3pt,belowskip=-1pt}
\centering
\begin{subfigure}{\textwidth}
\centering
\includegraphics[trim={0 0 0 0.39cm}, clip, width=0.7\textwidth]{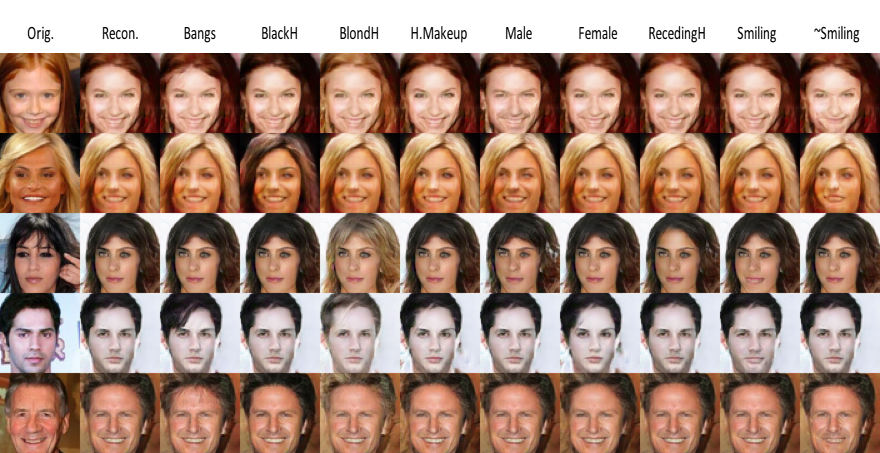}
\caption{BiCoGAN}
\label{fig:bcgan_efl_celeba_qual_autovar}
\end{subfigure}
\begin{subfigure}{\textwidth}
\centering
\includegraphics[trim={0 0 0 0.39cm}, clip, width=0.7\textwidth]{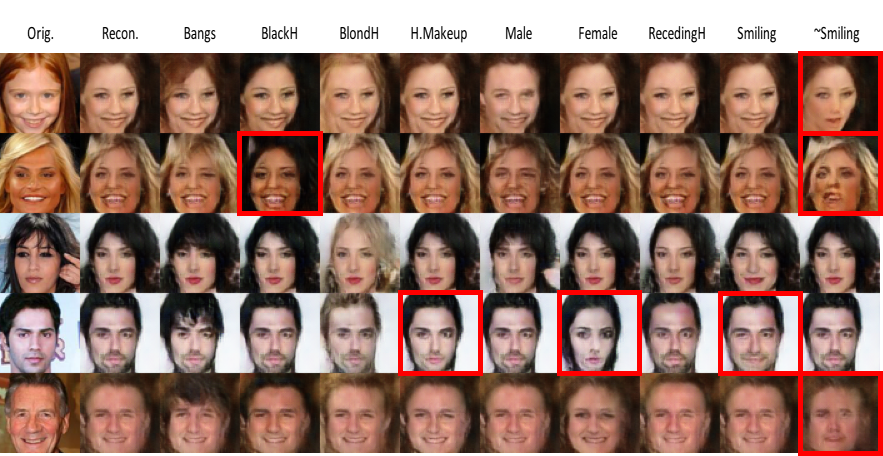}
\caption{IcGAN}
\label{fig:icgan_celeba_qual_autovar}
\end{subfigure}
\caption{CelebA images reconstructed with varied attributes. ``Orig'' shows the real image, ``Recon'' shows its reconstruction and the other columns show images with the same $z$ but varied $c$. Red boxes show cases where unspecified attributes or latent factors are \emph{mistakenly} modified during generation.}
\label{fig:bcgan_celeba_qual_autovar}
\end{figure}

We assess the performance of the generator and the encoder in the BiCoGAN framework jointly by comparing our model with IcGAN and cAVAE on the ability to reconstruct images with varied extrinsic factors on the MNIST dataset, and with IcGAN on the CelebA dataset. We do not compare against cALIM since it does not encode $c$. In order to vary $c$ while generating images with cAVAE, we first calculate the $s$-embedding for each class as we did in Section~\ref{sec:cond_gen}. Figures~\ref{fig:bcgan_mnist_qual_autovar}~and~~\ref{fig:bcgan_celeba_qual_autovar} show our results on MNIST and CelebA, respectively. We see that intrinsic factors (such as writing style for MNIST and subject identity, lighting and pose for CelebA) are better preserved in variations of images reconstructed with BiCoGANs compared to other models. On CelebA we also see that for BiCoGAN, changing an attribute has less effect on the incorporation of other extrinsic factors as well as the intrinsic features in the generation process, compared to IcGAN. This reinforces similar results that we observed in Section~\ref{sec:cond_gen}.

\begin{figure}
\captionsetup[subfigure]{font=scriptsize,labelfont=scriptsize,aboveskip=3pt,belowskip=-1pt}
\centering
\begin{subfigure}[c]{0.4\textwidth}
\centering
\includegraphics[trim={0 3.5cm 0 0}, clip, width=0.95\textwidth]{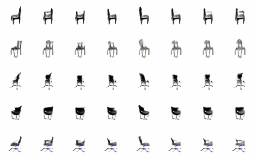}
\caption{Generated chairs}
\label{fig:bcgan_chairs_qual_rand}
\end{subfigure}
\begin{subfigure}[c]{0.59\textwidth}
\centering
\includegraphics[width=0.95\textwidth]{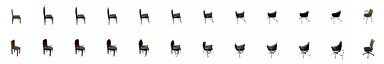}
\caption{Interpolation results}
\label{fig:bcgan_chairs_qual_interpolate}
\end{subfigure}
\caption{BiCoGAN results on Chairs dataset with continuous extrinsic attributes.}
\end{figure}

\subsection{Continuous Extrinsic Factors}

In previous subsections, we have provided results on datasets where $c$ is categorical or a vector of binary attributes. We evaluate the ability of the BiCoGAN to model data distributions when $c$ is continuous, on the Chairs dataset~\cite{bib:chairs} with $c$ denoting the yaw angle. Figure~\ref{fig:bcgan_chairs_qual_rand} shows chairs generated at eight different angles using our model, with $z$ fixed along rows. The results show that the model is able to generate chairs for different $c$ while preserving the information contained in $z$. We also assess the ability of BiCoGAN to learn the underlying manifold by interpolating between pairs of chairs. Figure~\ref{fig:bcgan_chairs_qual_interpolate} shows results of our experiments. Each row in the grid shows results of interpolation between the leftmost and the rightmost images. We see that the proposed BiCoGAN model shows smooth transitions while traversing the underlying latent space of chairs.

\begin{table}
\centering
\setlength\tabcolsep{5pt}
\caption{Accuracies of Predicting Additional Factors using Encoding - CelebA}
\label{tab:other_attr_pred}
\begin{tabular}{ c ? c | c ? c }
  \hbline
  \textbf{Attribute} & \textbf{LNet+ANet} & \textbf{WalkLearn} & \textbf{Ours} \\ \hbline
  5\_o\_Clock\_Shadow & 0.91 & 0.84 & \textbf{0.92} \\
  Arched\_Eyebrows & 0.79 & \textbf{0.87} & 0.79 \\
  Attractive & 0.81 & \textbf{0.84} & 0.79 \\
  Bags\_Under\_Eyes & 0.79 & \textbf{0.87} & 0.83 \\
  Big\_Lips & 0.68 & \textbf{0.78} & 0.70 \\
  Big\_Nose & 0.78 & \textbf{0.91} & 0.83 \\
  Blurry & 0.84 & 0.91 & \textbf{0.95} \\
  Chubby & 0.91 & 0.89 & \textbf{0.95} \\
  Double\_Chin & 0.92 & 0.93 & \textbf{0.96} \\
  Goatee & 0.95 & 0.92 & \textbf{0.96} \\
  High\_Cheekbones & 0.87 & \textbf{0.95} & 0.85 \\
  Narrow\_Eyes & 0.81 & 0.79 & \textbf{0.86} \\
  No\_Beard & \textbf{0.95} & 0.90 & 0.92 \\
  Oval\_Face & 0.66 & \textbf{0.79} & 0.74 \\
  Pointy\_Nose & 0.72 & \textbf{0.77} & 0.75 \\
  Rosy\_Cheeks & 0.90 & \textbf{0.96} & 0.94 \\
  Sideburns & \textbf{0.96} & 0.92 & \textbf{0.96} \\
  Wearing\_Earrings & 0.82 & \textbf{0.91} & 0.84 \\
  Wearing\_Lipstick & \textbf{0.93} & 0.92 & \textbf{0.93} \\
  Wearing\_Necklace & 0.71 & 0.77 & \textbf{0.86} \\
  Wearing\_Necktie & \textbf{0.93} & 0.84 & \textbf{0.93} \\
  Young & \textbf{0.87} & 0.86 & 0.85 \\ \hbline
  MEAN & 0.84 & \textbf{0.87} & \textbf{0.87} \\ \hbline
\end{tabular}
\end{table}

\subsection{Using The Learned Representation}

Finally, we quantitatively evaluate the encoding learned by the proposed BiCoGAN model on the CelebA dataset by using the inferred $z$ and $c$, i.e., the intrinsic factors and the $18$ extrinsic attributes on which the model is trained, to predict the other $22$ features annotated in the dataset. We train a simple feed-forward neural network for this task. Table~\ref{tab:other_attr_pred} shows the results of our experiment with the attribute-level breakdown of prediction accuracies. We show results of the state-of-the-art methods, LNet+ANet~\cite{bib:lnet_anet} and WalkLearn~\cite{bib:walk_learn}, for reference. The results show that it is possible to achieve state-of-the-art results on predicting these attributes by using the $z$ and $c$ encoded by the proposed BiCoGAN model, instead of original images. This not only shows that information about these attributes is captured in the encoded $z$ but also presents a successful use-case of the disentangled embedding learned by the BiCoGAN encoder.